\documentclass[sigconf]{acmart}

\AtBeginDocument{%
  \providecommand\BibTeX{{%
    \normalfont B\kern-0.5em{\scshape i\kern-0.25em b}\kern-0.8em\TeX}}}

\copyrightyear{2022}
\acmYear{2022}
\setcopyright{acmcopyright}\acmConference[WWW '22]{Proceedings of the ACM Web Conference 2022}{April 25--29, 2022}{Virtual Event, Lyon, France}
\acmBooktitle{Proceedings of the ACM Web Conference 2022 (WWW '22), April 25--29, 2022, Virtual Event, Lyon, France}
\acmPrice{15.00}
\acmDOI{10.1145/3485447.3512037}
\acmISBN{978-1-4503-9096-5/22/04}

\usepackage[utf8]{inputenc} 
\usepackage[T1]{fontenc}    
\usepackage{hyperref}       
\usepackage{booktabs}       
\usepackage{amsfonts}       
\usepackage{nicefrac}       
\usepackage{microtype}      
\usepackage{xcolor}         
\usepackage{multicol}

\usepackage{amsmath,wasysym}
\usepackage[ruled,vlined,linesnumbered, resetcount]{algorithm2e}

\usepackage{subcaption}
\usepackage{multirow}
\usepackage{textcomp}
\usepackage{footnote}
\usepackage{makecell}
\usepackage{xspace}
\usepackage{enumitem}
\usepackage{wrapfig}
\usepackage{amsthm}
\usepackage{wrapfig}

\usepackage[normalem]{ulem}
\useunder{\uline}{\ul}{}

\newcommand{\model}{\textsc{CAMul}\xspace}

\newtheorem{obs}{Obs}


\begin{document}

\title{CAMul: Calibrated and Accurate Multi-view Time-Series Forecasting}

\author{%
  Harshavardhan Kamarthi, Lingkai Kong, Alexander Rodr\'iguez, Chao Zhang, B. Aditya Prakash}
  \affiliation{
  College of Computing, 
  Georgia Institute of Technology
  \country{USA}}
  \email{{harsha.pk,lkkong,arodriguezc,chaozhang,badityap}@gatech.edu}

\begin{abstract}
Probabilistic time-series forecasting enables reliable decision making across many domains.
Most forecasting problems have diverse sources of data containing multiple modalities and structures.
Leveraging information from these data sources for
accurate and well-calibrated forecasts is an important but challenging problem.
Most previous works on multi-view time-series forecasting aggregate features from each data view by simple summation or concatenation and do not explicitly model uncertainty for each data view.
We propose a general probabilistic multi-view forecasting framework \model,
which  can learn  representations and uncertainty from diverse data sources. It integrates the information and uncertainty from each data view in a dynamic context-specific manner, assigning more importance to useful views to model a well-calibrated forecast distribution.
We use \model for multiple domains with varied sources and modalities and show that \model outperforms other state-of-art probabilistic forecasting models by over 25\% in accuracy and calibration.
\end{abstract}

 
\maketitle

\section{Introduction}
\label{sec:intro}

Time-series forecasting is a classic 
machine learning problem 
with applications covering wide-ranging domains including retail, meteorology, economics, epidemiology and energy. 
For many of these applications, we have a wide variety of datasets representing different \emph{views} 
or perspectives of the phenomena to forecast.
These views may differ in their structure and modality, and also in the quality and reliability due to collection and processing differences.
Due to the high-stakes decisions that these forecasts may inform (e.g., hospital resource allocation for COVID-19, planning energy infrastructure for cities), designing ML models that can leverage these multiple data sources to provide not only accurate but also well-calibrated forecast distributions is an important task.
\begin{figure}[tb]
    \centering
    \includegraphics[width=.85\linewidth]{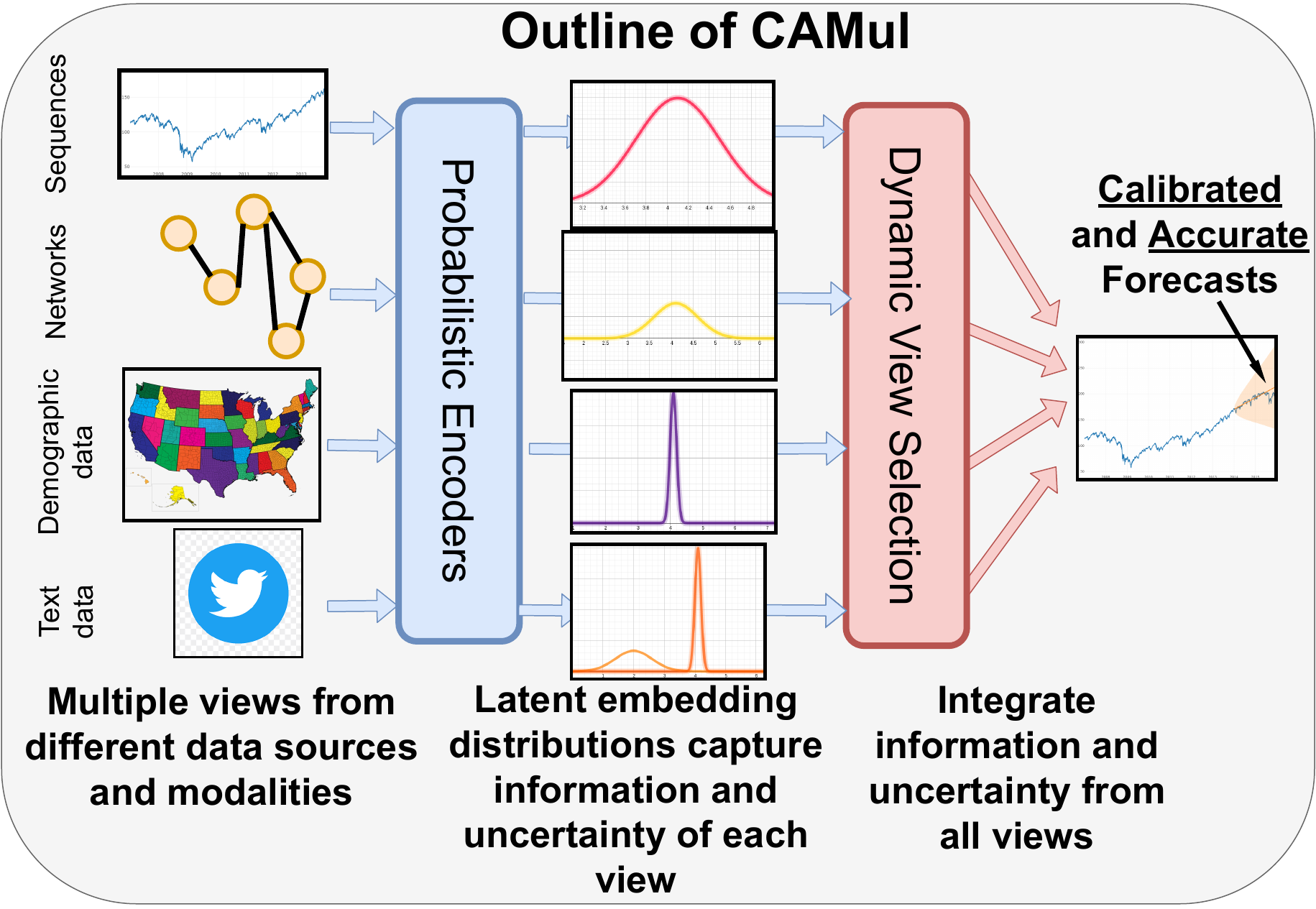}
    \vspace{-0.1in}
    \caption{\textit{\model models and integrates view-specific information and uncertainty.}}
    \vspace{-0.15in}
    \label{fig:teaser}
\end{figure}

State-of-the-art time series forecasting methods 
employ sequential neural models that can naturally integrate multiple time-varying exogenous features \cite{salinas2020deepar, li2021learning,rangapuram2018deep}. 
However, they are not suitable to ingest multiple modalities of data such as  networks, sequences and fixed sized features together.
Further, effective integration of information from multiple 
views
is challenging because information from individual views may be noisy, conflicting, redundant or sometimes unreliable.
Recent works that seek to integrate multi-source and multi-modal data for accurate sequence forecasting by using architectures like Graph Neural networks and Convolutional networks 
(such as for 
disease \cite{ramchandani2020deepcovidnet} and sales forecasting \cite{ekambaram2020attention}, multi-modal sentiment analysis \cite{wang2020transmodality}) do not comprehensively address this challenge. They propose to combine embeddings from multiple data sources either through direct concatenation \cite{hu2020understanding,zhong2020financial,trirat2021df} or simple summation \cite{ramchandani2020deepcovidnet}.
Other recent methods \cite{chen2021unite} learn a view-specific importance parameter and use it to combine embeddings in a weighted manner.
However, these weights are learned over all data points and are not
context-sensitive: they do not account for temporal variation in the
importance of each view and  variance across time series.



Moreover, 
these multi-modal time-series forecasting methods 
do not focus on learning a \emph{well-calibrated} forecast distribution.
This is especially challenging in a multi-view setting since the ambiguity and reliability of views has to be integrated to better inform forecast uncertainty.
Redundancy and high confidence in beliefs across multiple data views could inform a more confident forecast whereas conflicting information and lower confidence from some data sources may suggest higher uncertainty.
Past works on multi-view probabilistic forecasting \cite{yan2021deep} do not integrate stochastic information about beliefs from individual data views and instead use a deterministic embedding from fusion of view-specific embeddings to learn distribution parameters.
As a result, these methods do not model the view-specific uncertainty which may result in mis-calibrated forecasts.



Hence in this paper, our work tackles the challenge of modelling as well as integrating information and uncertainty from each data view to provide accurate and well-calibrated time-series forecasts (Figure \ref{fig:teaser}).
We introduce a general multi-view probabilistic time-series forecasting
framework, \model (\underline{C}alibrated and \underline{A}ccurate
\underline{Mul}ti-view Forecasting),  that jointly models uncertainty from
multiple views independently using a latent embedding distribution. It then
combines the views in a context-sensitive mechanism by accounting for their
reliability specific to the given sequence, thus providing well-calibrated and accurate forecasts. 
To learn a view-specific distribution capturing relevant information for each
data source, we use a non-parametric modelling framework. We directly use
latent embeddings of the data points in the training set of each view in the functional space to allow flexible representation of the predictive distribution that rely on similar patterns seen before. Our main contributions are:

    \noindent a) \textbf{Probabilistic Neural Framework jointly modeling multi-view uncertainty}: We propose a general framework \model for probabilistic time series forecasting on multi-modal and multi-source data
    making no assumptions on the structure of data.
    Our non-parametric probabilistic model leverages probabilistic relations learned between latent representations of data points for each data view to account for uncertainty from each view.
    
    \noindent b) \textbf{Integrating multi-view uncertainty towards calibrated forecasts}:
  \model leverages the latent information and uncertainty from each view and carefully integrates the beliefs from multiple views together, dynamically weighting each view's importance based on input data, to learn well-calibrated predictive distribution.
    
    \noindent b) \textbf{Evaluation of \model on multiple domains:} We use the \model framework to design models for multi-view time-series forecasting tasks from different domains using diverse  data sources and modalities (static features, sequences, networks).
    We compare \model against state-of-art domain-specific as well as general forecasting baselines and show that \model models clearly outperform all baselines by over 25\% in accuracy and calibration.
    We also show both empirically and using case studies, that our method of modeling and integrating uncertainty from individual data sources indeed causes these improvements.

\section{Related Work}

\noindent\textbf{Probabilistic Time-series Forecasting}
Classical time-series forecasting like exponential smoothing and ARIMA-based models \cite{hyndman2018forecasting} focus on univariate time-series with a small number of exogenous features and learn model parameters independently for each sequence. 
Recent probabilistic forecasting models leverage the representation power of neural sequential modules like
 DeepAR \cite{salinas2020deepar} which directly models the mean and variance parameters of the forecast distribution,  
 Bayesian Neural Networks \cite{zhu2017deep} which require assigning useful priors to parameters and require high computational resources for learning.
 Some recent works inspired from the space-state models explicitly model the transition and emission components with deep learning modules such as in the case of Deep Markov Models \cite{krishnan2017structured}, and recent Deep State Space models \cite{li2021learning, rangapuram2018deep}. Others introduce stochasticity into state dynamics of recurrent neural networks such as Stochastic RNN \cite{fraccaro2016sequential}, Variational RNN \cite{chung2015recurrent} and State Space LSTM \cite{zheng2017state}. 
Neural Process (NP)~\cite{garnelo2018neural} models a global latent variable for entire dataset to capture uncertainty with is used with input data's embedding to model the distribution parameters. Recurrent neural Process \cite{qin2019recurrent} leverages NP for sequence data. Functional Neural Process (FNP) captures stochastic correlations between input data and datapoints from training distribution to provide a flexible non-parametric mechanism to model output distribution using related training data points in functional space.
EpiFNP \cite{kamarthi2021doubt} a state-of-art calibrated disease forecasting model,  is closest to our work and leverages Functional Neural Process (FNP)~\cite{louizos2019functional}, which uses stochastic correlations between input data and datapoints to model a flexible non-parametric distribution for univariate sequences.
Our work leverages FNP for uncertainty modeling of each of the individual views before we jointly model the forecast distribution combining distributions from different views.

\noindent\textbf{Multi-view time-series forecasting}
Recent advances is deep learning architectures has allowed us to extract representations from variety of data sources such as images \cite{li2016survey}, sequences \cite{salehinejad2017recent}, graphs \cite{hamilton2020graph}, text \cite{devlin2018bert}, etc., and combine these modules' representation for training in a end-to-end fashion.
In order to integrate these representations, most methods employ simple summation or concatenation methods \cite{yan2021deep, li2018survey,zhong2020financial,trirat2021df,kumar2020edarkfind} either at the inital layers of model (early fusion) or last layers (late fusion) \cite{lahat2015multimodal,gadzicki2020early,luggen2021wiki2prop,wang2020transmodality}. For instance DeepCovidNet \cite{ramchandani2020deepcovidnet} uses spatio-temporal and static features using simple summation for Covid-19 prediction. \citet{ekambaram2020attention} similarly integrate images, text and static features for predicting consumer sales.
Moreover, most of these methods do not focus on probabilistic forecasting unlike \cite{chen2021unite} which uses  EHR sequence data, static demographic and location data and learn a data-source specific weight by pre-training the aggregated embeddings on prediction task of deciding if patient requires a treatment. Then, they employ a Gaussian Process approach to learn forecast probability on these embeddings to capture uncertainty.
In contrast, we  model data-view specific uncertainty by learning a latent distribution for each view which allows the model to reason about source specific uncertainty as it integrates uncertainty-aware stochastic representations of all views towards the forecast distribution.

\begin{figure*}[h]
    \centering
    \includegraphics[width=.92\linewidth]{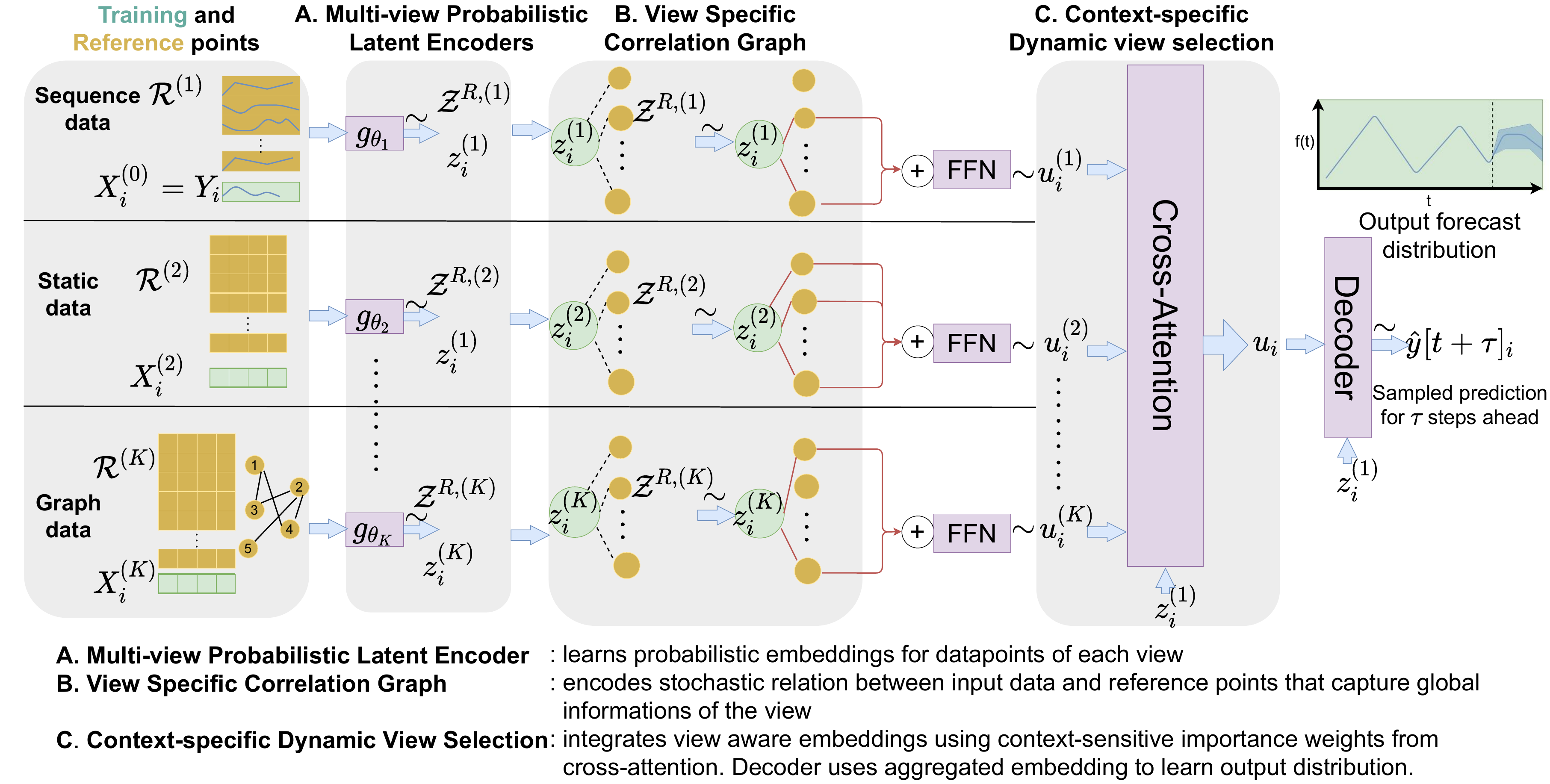}
    \vspace{-10pt}
    \caption{\textit{Pipeline of our method \model describing all the components of the generative process.}}
    \label{fig:pipeline}\vspace{-0.1in}
\end{figure*}

\section{Problem Formulation}

We describe the \emph{Multi-view probabilistic forecasting} problem, a generalization of the probabilistic forecasting problem to multi-modal and multi-source data.
We denote the time-series dataset as $\mathcal{Y} = \{Y_i\}_{i=1}^N$ where individual time-series is a univariate sequence, $Y_i = \{y[t]_i: y[t]_i \in \mathbf{R}, 1\leq t \leq T\}$. At each time-step $t$, we also have exogenous data sources that can be used to forecast future values in $\mathcal{Y}$. We divide these features from multiple sources and modalities into \emph{views}. Assume we have $K$ views for a specific problem representing $K$ sources/modalities of data. The dataset for a view $j$ is denoted by $\mathcal{X}^{(j)} = \{X_i^{(j)}\}_{i=0}^N$.
We do not assume any specific data type for each view. However, all datapoints of a view are assumed to have the same type. 
We can also have \emph{static views} whose values do not change across time: $  \forall t, x[t]_i^{(j)} = x[1]_i^{(j)}$. In some cases, we also denote as $X[1:t]_i^{(j)}$ all the data till week $t$: $X[1:t]_i^{(j)} = \{x[t']_i^{(j)}: 0\leq t' \leq t \}$.

Given the time-series values $\{Y[1:t]_i\}_{i=0}^N$ and exogenous features from all views $\{X[1:t]_i^{(j)}\}_{i=1\dots N, j=1\dots K}$ till current time $t$, our goal is to model a well-calibrated forecast probability distribution

\vspace{-0.1in}
{\small 
\[
    P( \{y[t+\tau]_i\}_{i=1}^N |  \{Y[1:t]_i\}_{i=1}^N, \{\{X[1:t]_i^{(j)}\}_{i=1\dots N}\}_{j=1\dots K})
\]}
for forecasting time-series values $\{y[t+\tau]_i\}_{i=1}^N$, $\tau$ steps in future.
We also assume that the view $1$ is the sequence dataset, i.e, $\forall i, Y_i = X_i^{(1)}$ and refer to it as the \emph{default view}.

\section{Methodology}
We first give an overview of \model and then describe each of its modules in more detail.
\par\noindent \textbf{Model Overview:} 
\model tackles the challenges related to jointly modeling useful information and uncertainty from multiple data-sources. 
It jointly models the predictive uncertainty by integrating information and uncertainty for each view. It dynamically considers the importance of each view based on input sequence to leverage the more important views and learn the output distribution.
We use the functional neural process (FNP) \cite{louizos2019functional} framework to learn a non-parametric latent embedding distribution for each view which enables a flexible method to model complex distribution by leveraging latent similarity with past training data-points from each data view.
 In contrast to related approaches that use FNP for univariate sequences \cite{kamarthi2021doubt},
\model solves a more general and harder problem of leveraging stochastic uncertainty from multi-source, multi-modal data and tackles the challenge of jointly modeling and integrating stochastic representations from each view.

At a high level, \model accomplishes our goals by performing the following steps in an end-to-end manner:
    A) learn latent representations for each view using \textit{Multi-view Latent Probabilistic Encoders} to capture information and uncertainty from each data source,
    B) leverage stochastic similarity between data-points in latent space to learn relations between time-series for each view  via \textit{View Specific Correlation Graph} and leverage these relations to derive a view-aware embedding distribution, 
    C) integrate information and uncertainty captured by latent embedding distributions from multiple views using learned importance of each view based on input sequence via \textit{Context-sensitive Dynamic View Selection Module} to derive the output distribution.
Figure \ref{fig:pipeline} shows the full pipeline.

For each view $j$, the set of data-points on which we learn the stochastic similarity w.r.t input point is a subset of the training set $\mathcal{X}^{(j)}$ of the view $j$ that comprehensively represents the entire training data. This set is called the \emph{reference set}, denoted by $\mathcal{R}^{(j)}$.
For example, for a view that represents sequence data from the past, we consider each of the complete time-series sequences  $X[1:t]_i^{(j)}$ from past training data as a single data-point in the reference set. 
Similarly consider views that represent static time-specific data such as months. Each training point for a given month contains the same static features. Therefore, the reference set contains a datapoint for each unique month.
Note that, during inference, we retain these reference sets and use them to derive the output distribution. 

\subsection{Multi-view Probabilistic Latent Encoders}
To capture useful information and uncertainty from each view, we learn a latent embedding distribution for datapoints of a specific view.
We leverage appropriate neural encoders for each view based on the modality to learn latent distribution parameters.
The probabilistic encoder $g_{\theta_j}$ for each view $j$ derives the parameters of latent embedding distribution for each of the reference points $\mathcal{R}^{(j)} = \{R_k^{(j)}\}_{k=1}^{N_j}$ and  input training point $X_i^{(j)}$ parameterized by a Gaussian Distribution:
\begin{equation}
\begin{split}
      \mu_k^{(j)}, \sigma_k^{(j)} = g_{\theta_j}(R_k^{(j)}),\quad\quad
    z_k^{(j)}  \sim \mathcal{N}(\mu_k^{(j)}, \sigma_k^{(j)}).
\end{split}
\label{eqn:latent}
\end{equation}
$g_{\theta_j}$, a neural network module parameterized by $\theta_j$, is chosen based on the modality of view $j$. 
For example, we use GRU for views with sequence data, Graph Convolutional Network \cite{Kipf2017SemiSupervisedCW} to encode relational data and Feed-Forward networks for static fixed-sized features.
The Supplementary contains detailed discussions of encoder architectures for different views used in this paper. The set of all latent encodings of reference set  $\mathcal{R}^{(j)}$ and training set $\mathcal{X}^{(j)}$ is denoted as $\mathcal{Z}^{R,(j)}$ and $\mathcal{Z}^{X,(j)}$ respectively.

\subsection{View Specific Correlation Graph}                  

The latent embedding for input data $z_i^{(j)} \in \mathcal{Z}^{X,(j)}$ captures the probabilistic information of input data in view $j$ whereas latent variables of $\mathcal{Z}^{R,(j)}$ capture stochastic information from entirety of the data view $j$. 
To capture the information from entire the data view $j$ conditioned on datapoint $i$, we use the non-parametric FNP framework that allows modeling a flexible data-view aware latent distribution for given input data by leveraging related reference points of the view. 

We first learn probabilistic relations between input datapoint and reference points in latent space using a similarity metric to create a \emph{View Specific Correlation Graph} (VSCG) and then use the reference points connected to input's embedding in VSCG to learn the \emph{view-aware latent embedding}.

We choose the radial basis function (RBF) as the similarity metric to relate reference set $\mathcal{R}^{(j)}$ to training set $\mathcal{X}^{(j)}$: 
\begin{equation}
    k(z_i^{(j)}, z_k^{(j)}) = \exp(-\rho ||z_k^{(j)}-z_i^{(j)}||^2).
\end{equation}
for all $z_k^{(j)} \in \mathcal{Z}^{R,(j)}$ and $z_i^{(j)} \in \mathcal{Z}^{X,(j)}$.

Next, we sample the VSCG $G^{(j)}$ as a bipartite graph between $\mathcal{R}^{(j)}$ and  $\mathcal{X}^{(j)}$, modeling each edge as a Bernoulli distribution parameterized by RBF similarity
\begin{equation}
    p((i,k)\in G^{(j)}) = \mathrm{Bern}(k(z_i^{(j)}, z_k^{(k)})).
\end{equation}
Note that during training, we approximate sampling from the discrete Bernoulli using a Gumbel-softmax distribution \cite{jang2016categorical}. 
Finally, we aggregate the sampled neighbouring reference points $N(i)$ for each training point $i$ to construct the \emph{view-aware latent variable}  $u_i^{(j)}$ (refer Figure \ref{fig:pipeline}) as:
\begin{equation}
\begin{split}
\mu(u_i^{(j)}), \hat{\sigma}(u_i^{(j)}) = \sum_{k\in N(i)} (l^{\mu}_j(z_k^{(j)}), l^{\sigma}_j(z_k^{(j)}))\\
    u_i^{(j)} \sim \mathcal{N}(\mu(u_i^{(j)}), \exp(\hat{\sigma}(u_i^{(j)}))),
\end{split}
\label{eqn:aggregate}
\end{equation}
where $l^{\mu}_j$ and $l^{\sigma}_j$ are linear layers. Thus, the view-aware embedding's latent distribution is derived from sampled reference points to capture  view-specific information. The stochastic process of VSCG also models view-specific uncertainty.

\subsection{Context-Specific Dynamic Views Selection}
Not all views are equally useful to learn a well-calibrated forecast distribution. The importance of each view for predictive distribution also varies for each time-series and across time.
 For instance, for  disease prediction, views representing features related to time (like months) may be useful for predicting seasonal changes but its importance may decrease during highly volatile and uncertain weeks near the peak where short-term sequence history and real-time exogenous features are more important. The importance of a view may also change dynamically when the features are corrupted for a small period of time during collection or measurement.

Thus, we need a dynamic mechanism to weigh the importance of each view based on the input sequence.
We propose the \emph{View Selection module} to learn importance weights for each view conditioned on input sequence and then aggregate embeddings from multiple views in proportion to learned weights.
Given the \emph{view-aware latent variables} of all $K$ views for each datapoint $i$ as  $\{u_i^{(j)}\}_{j=1}^K$, we combine these multiple views' knowledge towards the construction of final functional for predictive distribution. We leverage the cross-attention mechanism  \cite{vaswani2017attention} to learn the importance of each view but  also condition the weights on input sequence's representation from default view $1$ that encodes the time-series sequence $z_i^{(1)}$ as:
\begin{equation}
    \{\alpha_i^{(j)}\}_{j=1}^K = \text{Softmax}_j(\{h_1(z_i^{(1)})^T h_2(u_i^{(j)})\}_{j=1}^K),
    \label{eqn: attenview}
\end{equation}
where $h_1$ and $h_2$ are linear layers to transform both embeddings to same dimensions.
$\{\alpha_i^{(j)}\}_{j=1}^K$ denotes the importance of all views for sequence $i$. 
Finally, we use the weights to combine the latent representations for each view as
\begin{equation}
    \tilde{u}_i = \sum_{j=1}^K\alpha_i^{(j)} u_i^{(j)}.
    \label{eqn:svcgcombine}
\end{equation}
Thus, $\tilde{u}_i$, called the \emph{combined view embedding} represents combined knowledge from correlations learned by all views weighted by their importance for final prediction distribution. We denote the set of \emph{combined view embeddings} for all sequences as $\mathcal{U} = \{\tilde{u}_i\}_{i=1}^N$.

\subsection{Forecast Distribution Decoder}
Having extracted stochastic representations from each of the views leveraging latent encoders, VSCGs and the View selection module, we learn the output distribution via the \emph{Decoder module}. The decoder module uses the combined view embeddings $\mathcal{U}$ and the sequence embedding of default view $\mathcal{Z}^{X,(1)}$ to learn the parameters of output distribution. While $\mathcal{U}$ directly uses only the selectively aggregated embeddings of reference sets, $\mathcal{Z}^{X, (1)}$ leverages local historical patterns of input sequence including novel information that can be used to extrapolate beyond information from reference sets of multiple views.
The final decoder process is described as:
\begin{equation}
\begin{split}
    e_i & = z_i^{(1)} \oplus \tilde{u}_i\\
    \mu(y_i), \sigma_(y_i) & = d_1(e_i), \exp(d_2(e_i))\\
    \hat{y}_i^{(t+\tau)} & \sim \mathcal{N}(\mu(y_i), \sigma(y_i)),
\end{split}
\label{eqn:decoder}
\end{equation}
where $d_1$ and $d_2$ and feed-forward layers and $\oplus$ is the concatenation operator.
Here, we used a multivariate Gaussian distribution to parameterize $P(y_i^{(t+\tau)}| e_i)$ since our target sequence has real numbers. However, this framework can be extended to discrete and integer-valued output by carefully choosing the appropriate distribution and their relevant statistic to predict.

\subsection{Model Training and Inference}
The generative process of \model can be summarized as

{\small 
\[
\begin{split}
    & P(\{y_i^{t+\tau}\}_{i=1}^N| \mathcal{X}, \mathcal{R}) =
    \int \sum_{j=1}^{K} [ \underbrace{ P(\mathcal{Z}^{X, (j)}|\mathcal{X}^{(j)}) P(\mathcal{Z}^{R, (j)}|\mathcal{R}^{(j)})}_{\text{Stochastic latent encoders}}\\
    & \underbrace{P(G^{(j)}|\mathcal{Z}^{X, (j)}, \mathcal{Z}^{R, (j)}) P(\mathcal{U}^{(j)} | G^{(j)})}_{\text{VSCG}} ]\\
   &  \underbrace{P(\mathcal{U}|\{\mathcal{U}^{(j)}\}_{j=1}^K)}_{\text{View selection}} \underbrace{P(\{y_i^{t+\tau}\}_{i=1}^N| \mathcal{U}, \mathcal{Z}^{X, (1)})}_{\text{Output distribution}} d\mathcal\{{Z}^{(j)}\}_j d\mathcal{U}.
\end{split}
\]
}
Our objective is to increase the log-likelihood of above equation which is intractable due to integrals over real-valued random variables. Therefore, we use variational inference by approximating the posterior $\prod_{j=1}^K P(\mathcal{Z}^{(j)},\mathcal{U}^{(j)}, G^{(j)} | \mathcal{X}^{(j)}, \mathcal{R}^{(j)}) P(\mathcal{U} | \{\mathcal{U}^{(j)}\}_j)$ with the variational distribution :
\[
    \begin{split}
        q_j(\mathcal{U}^{(j)}, \mathcal{Z}^{(j)}, G^{(j)} | \mathcal{X}^{(j)} =  P(\mathcal{Z}^{X, (j)}|\mathcal{X}^{(j)}) P(\mathcal{Z}^{R, (j)}|\mathcal{R}^{(j)}) \\ P(G^{(j)}|\mathcal{Z}^{X, (j)}, \mathcal{Z}^{R, (j)} q_j(\mathcal{U}^{(j)} | \mathcal{X}^{(1)}),
    \end{split}
    \]
        
for each view $j$, where $q_j$ is a 2-layer network that parametrizes the Gaussian distribution $q_j(\mathcal{U}[j] | \mathcal{X}^{(1)})$ with input sequences $\mathcal{X}^{(1)}$. The ELBO is derived to be:

\[
\begin{split}
    \mathcal{L} &= -E_{q_j(\mathcal{Z}^{(j)}, G^{(j)} , \mathcal{U}^{(j)}| \mathcal{X}^{(j)}) } [\log P(\{y_i^{t+\tau}\}_{i=1}^N| \mathcal{U}, \mathcal{Z}^{X, (1)})\\
    &+ \sum_{j=1}^K \log P(\mathcal{U}^{(j)} | G^{(j)}, \mathcal{Z}^{(j)}) - \log q_j(\mathcal{U}^{(j)} | \mathcal{X}^{(1)})].
\end{split}
\]

The parameters of the inference distributions $q_j$ and the components of the generative process are jointly learned via Stochastic Gradient Variational Bayes to minimize the ELBO loss \cite{kingma2013auto}. We use the reparametrization trick for all sampling processes. The pseudo-code for training is available in Appendix.
During inference, we sample from the joint distribution $P(\{y_i^{t+\tau}\}_{i=1}^N, \{\mathcal{Z}^{(j)},\\ G^{(j)}\}_j, \mathcal{U}| \mathcal{X}, \mathcal{R})$
to generate samples for the forecast distribution.

\section{Experiments}

\label{sec:expt}
 We evaluated our models on a workstation that runs on Intel Xeon 64 core processor with 128 GB memory on a Nvidia Tesla V100 GPU. Our model takes less than 6 GB of memory and takes 20-40 minutes of training time for each of the benchmarks.
 We have released the code and datasets publicly \footnote{\url{https://github.com/AdityaLab/CAMul}} and describe in detail the hyperparameters in Appendix.
 Next, we describe the baselines, benchmark datasets and tasks as well as the evaluation metrics.
 \vspace{-0.1in}
 \subsection{Setup}
 \label{sec:setup}
 \subsubsection{\textbf{Baselines}}
We compare our model against the state-of-art probabilistic forecasting baselines along with some domain-specific baselines. 
The chosen forecasting baselines have shown state-of-art performance on a wide set of probabilistic forecasting tasks such as power consumption, air quality, traffic forecasting, health risk assessment, data center load estimation, etc.
    \noindent$\bullet$ \textbf{SARIMA} \cite{hyndman2018forecasting}: A classic time-series forecasting baseline based on ARIMA that accounts for seasonal shifts.
     $\bullet$ \textbf{DeepAR} \cite{salinas2020deepar}: A state-of-art, widely used RNN based probabilistic forecasting model that learns a parametric distribution.
     $\bullet$ \textbf{Deep State Space Model (DSSM)} \cite{li2021learning}: A space state based model that uses neural networks to model transition and emission distribution of state space.
     $\bullet$ \textbf{Deep Graph Factors (GraphDF)} \cite{chen2021graph}: A deep probabilistic forecasting model that integrates relational information from graphs across time-series. In absence of explicit relational information, \cite{chen2021graph} proposes to use RBF kernel over the euclidean distance as edge weights over pairs of time-series, which we use as the \textbf{GraphDF-RBF} baseline. If explicit relational data in form of adjacency graph is available (in case of all benchmarks except \texttt{power}), we also evaluate using this adjacency graph as the \textbf{GraphDF-Adj} baseline.
     $\bullet$ \textbf{Recurrent neural process (RNP)} \cite{qin2019recurrent}: Neural Process based method for temporal data which uses attention over reference points as part of the generative process.
     $\bullet$ \textbf{Multi-modal Gaussian Process (MMGP)}: Similar to \cite{chen2021unite} we first pre-train a deterministic model combining deterministic embeddings from the encoders and aggregating embeddings with fixed learned weights to predict the output. Then we aggregate the embeddings as features to train a Gaussian Process to get probabilistic predictions. 

For disease forecasting benchmarks, we also evaluate against state-of-art domain-specific forecasting models that use the same set of exogenous features as \model. 
We choose top performing deep learning models for flu-forecasting \texttt{google-symptoms} task: \textbf{EpiDeep} \cite{adhikari2019epideep} and \textbf{EpiFNP} \cite{kamarthi2021doubt}. For the \texttt{covid19} task, we also choose \textbf{CMU-TS} \cite{cramer2021evaluation1}  and \textbf{DeepCovid} \cite{rodriguez2020deepcovid} as baselines; these leverage the CovDS dataset \cite{kamarthi2021back2future} and are top performing statistical models at the Covid-19 Forecast Hub organized by CDC \cite{cramer2021evaluation1}.
We also evaluate against variants of \model which enable us to examine the importance of Context-Specific View Selection and utility of our view-specific  probabilistic modeling approach over the widely used method of probabilistic modeling on the fused deterministic embedding of view-specific representations (see Q2 of Section \ref{sec:results}).
\vspace{-0.1in}
 
\subsubsection{\textbf{Benchmarks}}
\label{sec:tasks}
We evaluate \model framework on time-series forecasting problem from a variety of domains involving diverse data views. For each benchmark, we describe the datasets used and the views corresponding to the features used from datasets.

\label{sec:bench}

\noindent\textbf{1. \texttt{google-symptoms} (Flu forecasting from Google symptoms):}
We use aggregated and anonymized search counts of flu-related symptoms web searches by Google\footnote{\url{https://pair-code.github.io/covid19_symptom_dataset}} from each US state to predict incidence of influenza from 1 to 4 weeks ahead in future. 

\noindent\underline{Dataset:}
The flu incidence rate is represented by wILI (weighted Influenza related illness) values released weekly by CDC for 8 HHS regions of USA\footnote{\url{https://predict.cdc.gov/post/5d8257befba2091084d47b4c}}. The aggregate symptom counts for over 400 symptoms are anonymized and publicly released for each week since 2017 at county and state level by Google \cite{googledataset}. We choose to extract 14 symptoms related to influenza referred to in the CDC website\footnote{\url{https://www.cdc.gov/flu/symptoms/symptoms.htm}}.
We aggregate these counts for each HHS region and use it as exogenous features. We also use spatial adjacency data between HHS regions and time-related data (month and year) as features.
For forecasting wILI values for a given year, we use the training set from all past years for model training and hyperparameter tuning.\\
    \noindent\underline{Views:} We use following diverse views for this benchmarks:
    1. \textit{Default Historical wILI view}: This view contains reference points of symptoms features for all previous forecasting seasons for all HHS regions. We use the GRU encoder for this view.
    2.  \textit{HHS adjacency view}: This is a graph feature view. We have 8 reference points corresponding to HHS regions. We aptly use the GCN based latent encoder where the features $\mathcal{F}^{(j)}$ are one-hot encodings and the adjacency matrix $A^{(j)}$ encodes the neighbourhood information between HHS regions that share a border. 
    3. \textit{Month view}: This is a static feature view. We have 12 reference points for each month. We use an embedding layer to learn encodings for each  reference point from one-hot embedding and use the feed-forward latent encoder.



\noindent\textbf{2. \texttt{covid19} (Covid-19 mortality forecasting):}
We evaluate our model on the challenging task of forecasting COVID-19 mortality for each of the 50 US states. 

\noindent\underline{Dataset:} We use the mortality data and exogenous features of CovDS data used in \cite{rodriguez2020deepcovid, kamarthi2021back2future}. The exogenous features contain multiple kinds of data including hospital records, mobility, exposure and web-based survey data. We evaluate for duration of 7 months from June 2020 to December 2020 and again use only past weeks' data to train before forecasting for 1 to 4 weeks ahead mortality.
We follow the real-time forecasting setup of \cite{rodriguez2020deepcovid} and train the model separately for each week over all states using data available up to past week as training set.

\noindent\underline{Views:} 1. \textit{Default Mortality view}: This is a sequence view where each reference point is a time-series of past mortality.  2. \textit{Line-list view} This is a multivariate sequence view that contains 7 weekly-collected features from traditional surveillance including hospitalization and testing. 3. \textit{Mobility and exposure view} Similarly we use the digitally collected signals measuring aggregate mobility and individuals collected from smartphones. For views 1,2 and 3 we use a GRU based encoder. 4. \textit{Demographic view} This is a static view where we encode each of the 50 states using 8 demographic features including average income, unemployment, population, average age and voting preferences.  We use a feed-forward network for encoder. 5. \textit{State adjacency view}: We construct a graph view of 50 states and encode spatial adjacency states. We use a GCN-based encoder.


\noindent\textbf{3. \texttt{power} (Power consumption forecasting):}
We evaluate on power consumption data which is a standard forecasting benchmark.

\noindent\underline{Dataset:} The dataset~\cite{powerdataset} contains 260,640 measurements of power consumption of a household for a year using measurements from 3 meters with a total of 7 features. Our goal is to forecast the total active power consumption for 1 minute in future. We use the data from the first 9 months of measurement as training set and last 3 months as test set for forecasting.

\noindent\underline{Views:} 1. \textit{Default Past sequence view} We randomly sample single day sequence from each of the past months as reference sets.  2. \textit{Month view}: Similar to Month view for \texttt{google-symptoms} benchmark. 3. \textit{Time of Day view}: This is a static view. We divide the 24 hours of a day into 6 equal intervals and assign each of 6 reference point an interval. Similar to Month view we use a embedding layer on one-hod encodings to represent each reference point.

\noindent\textbf{4. \texttt{tweet} (Tweet Topics prediction):}
The goal for this task is to evaluate the topic distribution of tweets in the future given topic distributions of past weeks similar to \citet{shi2019state}.

\noindent\underline{Dataset:} We collect Covid-19 related tweets for 15 weeks that have a geographical tag to identify US state of the user. From the tweet text, we extract 30 topics using LDA \cite{blei2003latent} and allocate each tweet from a given week to each of the 30 topics it is most likely related for each state. Thus, we have a multivariate sequence dataset where for each of the 50 states, we have sequences containing topic distribution of tweets for each week. Similar to \texttt{google-symptoms}, for each forecasting for each year, we use data from past year as training set. We train a separate decoder module for each of the 30 topics and report the scores averaged over all topics.

\noindent\underline{Views:} 1. \textit{Default Past sequence view}: Similar to other task the first view contains the past sequences for each state. 2. \textit{State adjacency view}: We construct a state adjacency view similar to \texttt{covid19} task. 3. \textit{Month view}: Similar to Month view of \texttt{power} and \texttt{google-symptoms}. 4. \textit{Demographics view}: Similar to Demographic view of \texttt{covid19}.
\vspace{-0.1in}

\subsubsection{\textbf{Evaluation metrics}}
Given the large set of evaluation metrics proposed for both point-prediction and probabilistic predictions \cite{hyndman2006another,tsyplakov2013evaluation,jordan2017evaluating}, we evaluate our model and baselines using carefully chosen metrics that are widely used in machine learning literature, relevant to our benchmarks and comprehensively evaluate both accuracy and calibration of models' forecasts.

\noindent\textbf{Root Mean Squared Error (RMSE)} is a popular metric used to evaluate accuracy of point prediction and is a better measure of robustness of model over metrics like MAE or MAPE since it is sensitive to instances of large errors.

\noindent\textbf{Interval Score (IS)} is a standard score used in evaluation of accuracy of probabilistic forecasts in epidemiology \cite{reich2019collaborative}. IS measures the negative log likelihood of a fixed size interval around the ground truth under the predictive distribution: 
\vspace{-0.05in}
\[IS(\hat{p}_y,y) = - \int_{y-L}^{y+L} \log \hat{p}_y(\hat{y}) d\hat{y}.\]
\noindent\textbf{Confidence Score (CS)}  introduced in \cite{kamarthi2021doubt} measures the overall calibration of predictions of a model $M$ similar to \cite{kuleshov2018accurate}. We first calculate the fraction $k_m(c)$ of prediction distributions that cover the ground truth at each confidence interval $c$. A perfectly calibrated model has $k_m(c)$ very close to $c$. Therefore, $CS$ is defined as:
\[CS(M) = \int_0^1 |k_m(c)-c| dc.\]
which is approximated by summation over small intervals \cite{kamarthi2021doubt}.

\noindent\textbf{Cumulative Ranked Probability Score (CRPS)} is a widely used standard metric for evaluation of probabilistic forecasts that generalizes mean average error to probabilistic forecasting \cite{gneiting2014probabilistic}. Since it is a proper scoring rule that is minimized if the prediction distribution matches, on expectation, the true distribution, CRPS measures both accuracy and calibration.
Given ground truth $y$ and the predicted probability distribution $\hat{p}(Y)$, let $\hat{F}_y$ be the CDF. Then, CRPS is defined as:
\[CRPS(\hat{F}_y, y) = \int_{-\infty}^\infty (\hat{F}_y(\hat{y}) - \mathbf{1}\{\hat{y}>y\})^2 d\hat{y}.\]

\begin{table*}[!th]

\scalebox{0.85}{
\begin{tabular}{lrrrr|lrrrr}
\hline
               & \multicolumn{4}{c|}{\texttt{covid19}}                                                                       & \multicolumn{5}{c}{\texttt{google-symptoms}}                                                                                \\
               & \multicolumn{1}{c}{RMSE}                & \multicolumn{1}{c}{CRPS} & \multicolumn{1}{c}{CS} & \multicolumn{1}{c|}{IS} &                & \multicolumn{1}{c}{RMSE}                 & \multicolumn{1}{c}{CRPS} & \multicolumn{1}{c}{CS} & \multicolumn{1}{c}{IS} \\ \hline
SARIMA         & 91.3$\pm$3.7   & 69.1$\pm$4.2   & 0.41$\pm$0.015 & 6.67$\pm$0.034            & SARIMA         & 1.72$\pm$0.021          & 1.62$\pm$0.016              & 0.4$\pm$0.005             & 5.04$\pm$0.32             \\
DeepAR         & 49.1$\pm$6.9   & 48.5 $\pm$ 7.1 & 0.19$\pm$0.015 & 3.72$\pm$ 0.043           & DeepAR         & 0.68$\pm$0.024          & 0.97$\pm$ 0.021             & 0.15$\pm$0.006            & 2.89$\pm$0.21             \\
DSSM           & 54.7$\pm$5.7   & 58.6$\pm$3.5   & 0.19$\pm$0.006 & 4.13$\pm$0.013            & DSSM           & 0.78$\pm$0.026          & 1.03$\pm$0.015              & 0.15$\pm$0.007            & 3.08$\pm$0.16             \\
RNP            & 52.9$\pm$5.3   & 71.8$\pm$6.5   & 0.32$\pm$0.015 & 5.31$\pm$0.042            & RNP            & 0.82$\pm$0.043          & 0.95$\pm$0.035              & 0.41$\pm$0.004            & 3.87$\pm$0.18             \\
MMGP             & 48.7$\pm$4.2   & 42.3$\pm$2.4   & 0.23$\pm$0.015 & 4.27$\pm$0.022            & MMGP             & 1.24$\pm$0.077          & 1.02$\pm$0.031              & 0.26$\pm$0.003            & 2.28$\pm$0.42             \\
GraphDF-RBF    & 51.3$\pm$3.7   & 43.2$\pm$2.1   & 0.19$\pm$0.013 & 3.41$\pm$0.035            & GraphDF- RBF   & 0.73$\pm$0.031          & 1.05$\pm$0.019              & 0.11$\pm$0.003            & 1.83$\pm$0.27             \\
GraphDF-Adj    & 48.5$\pm$4.2   & 49.1$\pm$3.4   & 0.23$\pm$0.021 & 3.79$\pm$0.023            & GraphDF- Adj   & 0.91$\pm$0.027          & 1.10$\pm$0.042              & 0.12$\pm$0.005            & 2.64$\pm$0.53             \\ \hline
DeepCovid      &38.3$\pm$5.2   & 39.8$\pm$6.6   & 0.22$\pm$0.013 & 3.63$\pm$0.046            & EpiDeep        & 0.98$\pm$0.053          & 1.07$\pm$0.062              & 0.29$\pm$0.006            & 5.55$\pm$0.84             \\
CMU-TS         & 32.4$\pm$5.3   & 30.1$\pm$5.2   & 0.21$\pm$0.014 & 3.31$\pm$0.062            & EpiFNP         & 0.64$\pm$0.048          & 0.52$\pm$0.057              & 0.05$\pm$0.004            & 0.67$\pm$0.12             \\ \hline
\textbf{\model} &\textbf{27.3 $\pm$ 3.5} & \textbf{23.8$\pm$4.1} & \textbf{0.14$\pm$0.011} & \textbf{2.08$\pm$0.015}   & \textbf{\model} & \textbf{0.49$\pm$0.072} & \textbf{0.34$\pm$0.053}     & \textbf{0.04$\pm$0.006}   & \textbf{0.54$\pm$0.05}    \\ \hline
\model-C        & 34.2$\pm$-4.2           & 27.7$\pm$3.2          & 0.18$\pm$0.006          & 2.83$\pm$0.025            & \model-C        & 0.53$\pm$0.048          & 0.44$\pm$0.031              & 0.06$\pm$0.008            & 0.6$\pm$0.07              \\
\model-S        & 31.7$\pm$4.6            & 26.8$\pm$3.1          & 0.17$\pm$0.005          & 2.75$\pm$0.042            & \model-S        & 0.62$\pm$0.082          & 0.42$\pm$0.058              & 0.06$\pm$0.003            & 0.6$\pm$0.03              \\
\model-D        & 43.8 $\pm$5.3           & 39.2$\pm$6.1          & 0.28$\pm$0.012          & 4.06$\pm$0.032             & \model-D        & 1.32$\pm$0.067          & 0.99$\pm$0.072              & 0.19$\pm$0.005            & 2.24$\pm$0.05            \\\hline
               & \multicolumn{4}{c|}{\texttt{power}}                                                                        & \multicolumn{5}{c}{\texttt{tweet}}                                                                                        \\
               & \multicolumn{1}{c}{RMSE}                & \multicolumn{1}{c}{CRPS} & \multicolumn{1}{c}{CS} & \multicolumn{1}{c|}{IS} &                & \multicolumn{1}{c}{RMSE}                 & \multicolumn{1}{c}{CRPS} & \multicolumn{1}{c}{CS} & \multicolumn{1}{c}{IS} \\ \hline
SARIMA         & 1.51$\pm$0.021              & 1.16 $\pm$ 0.018        & 0.23 $\pm$ 0.002          & 3.49 $\pm$0.021           & SARIMA      & 0.33$\pm$0.031              & 1.41 $\pm$ 0.262          & 0.37$\pm$ 0.007           & 3.87 $\pm$ 0.74          \\
DeepAR         & 1.01$\pm$0.008              & 0.72 $\pm$ 0.002        & 0.04 $\pm$ 0.003          & 1.57 $\pm$ 0.046          & DeepAR      & 0.11$\pm$0.054              & 1.06 $\pm$ 0.152          & 0.15 $\pm$ 0.003          & 1.35 $\pm$ 0.31          \\
DSSM           & 0.91$\pm$0.024              & 0.59 $\pm$ 0.018        & \textbf{0.02 $\pm$ 0.002} & 1.32 $\pm$ 0.054          & DSSM        & 0.08$\pm$0.036              & 0.92 $\pm$ 0.173          & 0.07 $\pm$ 0.002          & 0.96 $\pm$ 0.34          \\
RNP            & 1.11$\pm$0.014              & 0.85 $\pm$ 0.015        & 0.19 $\pm$ 0.005          & 3.31 $\pm$ 0.018          & RNP         & 0.11$\pm$0.065              & 1.16 $\pm$ 0.045          & 0.22 $\pm$ 0.006          & 3. 05 $\pm$ 0.52         \\
MMGP           & 1.28$\pm$0.036              & 1.19 $\pm$ 0.041        & 0.15 $\pm$ 0.001          & 3.03 $\pm$ 0.013          & MMGP        & 0.21$\pm$0.052              & 1.31 $\pm$ 0.162          & 0.13 $\pm$ 0.0026         & 1.65 $\pm$ 0.49          \\
GraphDF-RBF    & 0.94$\pm$0.031              & 0.92 $\pm$ 0.03         & 0.13 $\pm$ 0.004          & 1.45 $\pm$ 0.033          & GraphDF-RBF & 0.12$\pm$0.073              & 0.62 $\pm$ 0.052          & 0.08 $\pm$ 0.010          & 1.91 $\pm$ 0.76          \\
               & \multicolumn{1}{l}{}     & \multicolumn{1}{l}{} & \multicolumn{1}{l}{}   & \multicolumn{1}{l|}{}  & GraphDF-Adj & 0.18$\pm$0.025              & 0.91 $\pm$ 0.031          & 0.14$\pm$ 0.009           & 2.39 $\pm$ 0.26          \\ \hline
\textbf{\model} & \textbf{0.85$\pm$0.027}              & \textbf{0.46 $\pm$0.02} & \textbf{0.02 $\pm$ 0.001} & \textbf{0.93 $\pm$ 0.021} & \model       & \textbf{0.07$\pm$0.005}     & \textbf{0.42 $\pm$ 0.015} & \textbf{0.05 $\pm$ 0.004} & \textbf{0.68 $\pm$ 0.21} \\ \hline
\model-C        & 1.03$\pm$0.035              & 1.13$\pm$0.04           & 0.08$\pm$0.001            & 2.88$\pm$ 0.014           & \model-C     & 0.08$\pm$0.053              & 0.84$\pm$0.027            & 0.07$\pm$0.002            & 1.03$\pm$0.17            \\
\model-S        & 0.99$\pm$0.036              & 0.68$\pm$0.05           & 0.04$\pm$0.002            & 1.27$\pm$0.027            & \model-S     & 0.19$\pm$0.048              & 0.58$\pm$0.081            & 0.05$\pm$0.006            & 1.41$\pm$0.25            \\
\model-D        & 1.31$\pm$0.057              & 1.36$\pm$0.07           & 0.14$\pm$0.001            & 2.96$\pm$0.012            & \model-D     & 0.18$\pm$0.077              & 1.15$\pm$0.020            & 0.19$\pm$0.001            & 1.14$\pm$0.15 \\\hline           
\end{tabular}
}

\caption{\textit{Evaluation scores (over 20 runs) for \model and baselines for all benchmarks.
We performed t-test with $\alpha=1\%$. Best scores are in \textbf{bold} and are statistically significantly better than other models. \model consistently performs the best with over 24\% improvement in accuracy and over 35\% improvement in calibration over the best baselines.}}
\label{tab:metrics}
\vspace{-0.3in}
\end{table*}

\vspace{-0.1in}
\subsection{Results}
\label{sec:results}

\textbf{Q1}: \textit{Does \model provide well-calibrated accurate probabilistic forecasts across all benchmarks?}

We evaluate our model and baselines on the four diverse benchmarks described in Section \ref{sec:setup}. We ran the experiments 20 times for each of the models for all tasks and reported the mean scores. Specifically, for \texttt{covid19} and \texttt{google-symptoms} we performed a comprehensive evaluation across all regions in US for 1-4 weeks ahead forecasts and reported the average results. The results are summarized in Tables \ref{tab:metrics}. 
\model models significantly outperform the baselines in both accuracy and calibration scores. Specifically for the hard disease-forecasting tasks we consistently see over 25\% improvement in RMSE and CRPS score, and over 50\% improvement in interval score over best baselines. In case of \texttt{power} and \texttt{tweet}, we observe 28\% and 24\% improvement in CRPS scores over second-best model and 56\% and 35\% improvement in interval score. Performing significance test (t-test) with $\alpha=1\%$ shows that
\model is significantly better than other models in all scores except those highlighted in bold in Table \ref{tab:metrics}. 
Further, applying post-hoc calibration methods \cite{kuleshov2018accurate, song2019distribution} on baselines also does not affect the significance of our results (Table \ref{tab:posthoc} in Supplementary).

\noindent\textbf{Q2} \textit{: Effect of  multi-source stochastic modelling and context-sensitive dynamic view selection on \model's performance}


We evaluate the efficacy  of 1) Context-Specific Dynamic Views Selection and 2) probabilistic modeling of each data view via Multi-view Latent Probabilistic Encoders and VSCG.
We compare the attention based dynamic view selection of \model with two other variants a) \textbf{\model-C} \textit{concatenates} VSCG-based latent embeddings from all views, 
b) \textbf{\model-S} learns a \textit{static} weight $w^{(j)}$ for each view $j$ and combines the VSCG-based embeddings: $\Tilde{u}_i = \sum_j w^{(j)} u_i^{(j)}$, 
c) To test the efficacy of stochastic modelling of latent embeddings, the variant \textbf{\model-D} uses \textit{deterministic} view-specific latent variables:
We use the latent encoders' mean rather than sampling from Gaussian as Eqn. \ref{eqn:latent} and use a cross-attention layer over reference points instead of VSCG to derive the view specific latent variables as a weighted summation of reference points.
The evaluation scores of the variants are shown in Table~\ref{tab:metrics}.
We see that the original configuration of \model with dynamic view selection and multi-source probabilistic modeling outperforms the variants in all benchmarks. Using deterministic multi-source latent embeddings, in particular, drastically decreases the scores emphasizing the efficacy of uncertainty modeling.

\noindent\textbf{Q3}: \textit{Does \model handle information and uncertainty from multiple views to provide better performance?}
\vspace{-0.2in}
\begin{table}[h]
\scalebox{0.8}{
\begin{tabular}{lcccc}
              & \texttt{power}                & \texttt{tweet}                  & \texttt{covid19}               & \texttt{google-symptoms}        \\ \hline
All Views     & \textbf{0.46} & \textbf{0.42 } & \textbf{27.3 } & \textbf{0.34 } \\
Default view  & 1.03          & 0.95            & 37.2            & 0.58            \\
Two Best Views        & 0.63          & 0.77            & 33.1            & 0.46            \\
Best Baseline & 0.59          & 0.62            & 32.4            & 0.64           
\end{tabular}
}
\caption{\textit{Comparison of CRPS score of \model with all views, the default view, two best views and the best baseline.}}
\label{tab:ablation}
\vspace{-0.3in}
\end{table}

\model models useful patterns and uncertainty for each of the diverse set of data views independently before combining them. This is in contrast to most baselines that use simple aggregation at feature level or latent embedding level where noisy or reliable data from some views may hinder performance. \model however learns to weigh the importance of each view before combining them for prediction.
To test the efficacy of \model's handling of multiple data views, we evaluated the model with a single default view, two best views and compared it with the original \model with all views and the best performing baseline as shown in Table \ref{tab:ablation}. 
Note that for all benchmarks, the best view is always the default view.
We see that model with all views is clearly the best performing. Moreover, using only the best view sometimes leads to lower performance compared to the best performing baseline which has access to data from all views. Therefore, we conclude that \emph{\model can handle multiple views that in turn lead to significantly better performance}.

\noindent\textbf{Q4}: \textit{Do the weights of each view from View Selection module correspond to predictive utility of the view?}
\vspace{-.1in}
\begin{figure}[h]
    \centering
    \begin{subfigure}{.47\linewidth}
    \centering
    \includegraphics[width=.98\linewidth]{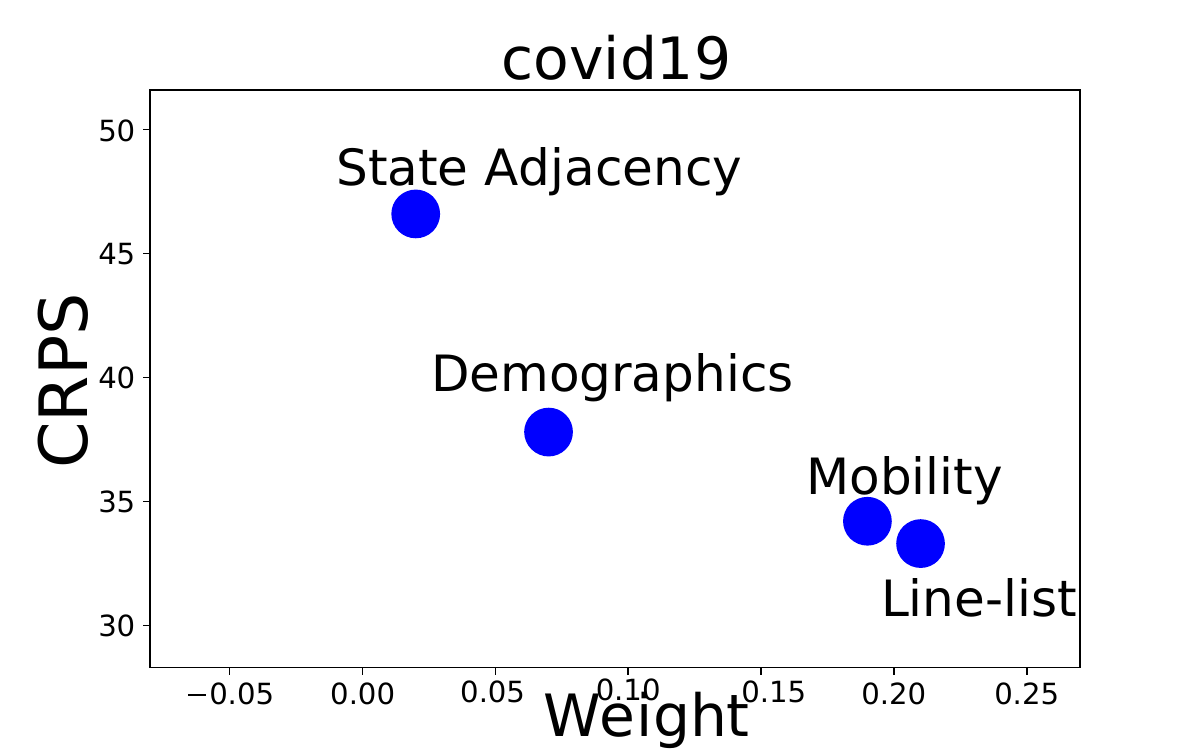}
    \end{subfigure}%
    \begin{subfigure}{.47\linewidth}
    \centering
    \includegraphics[width=.98\linewidth]{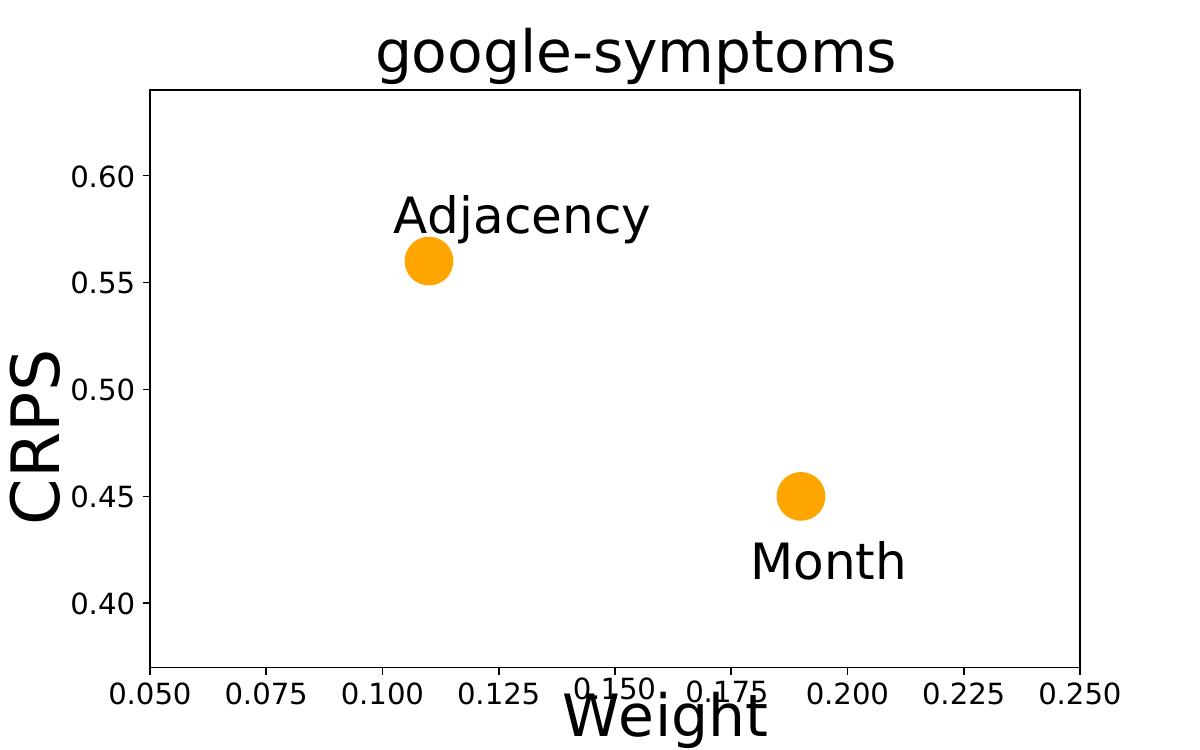}
    \end{subfigure}
    \begin{subfigure}{.47\linewidth}
    \centering
    \includegraphics[width=.98\linewidth]{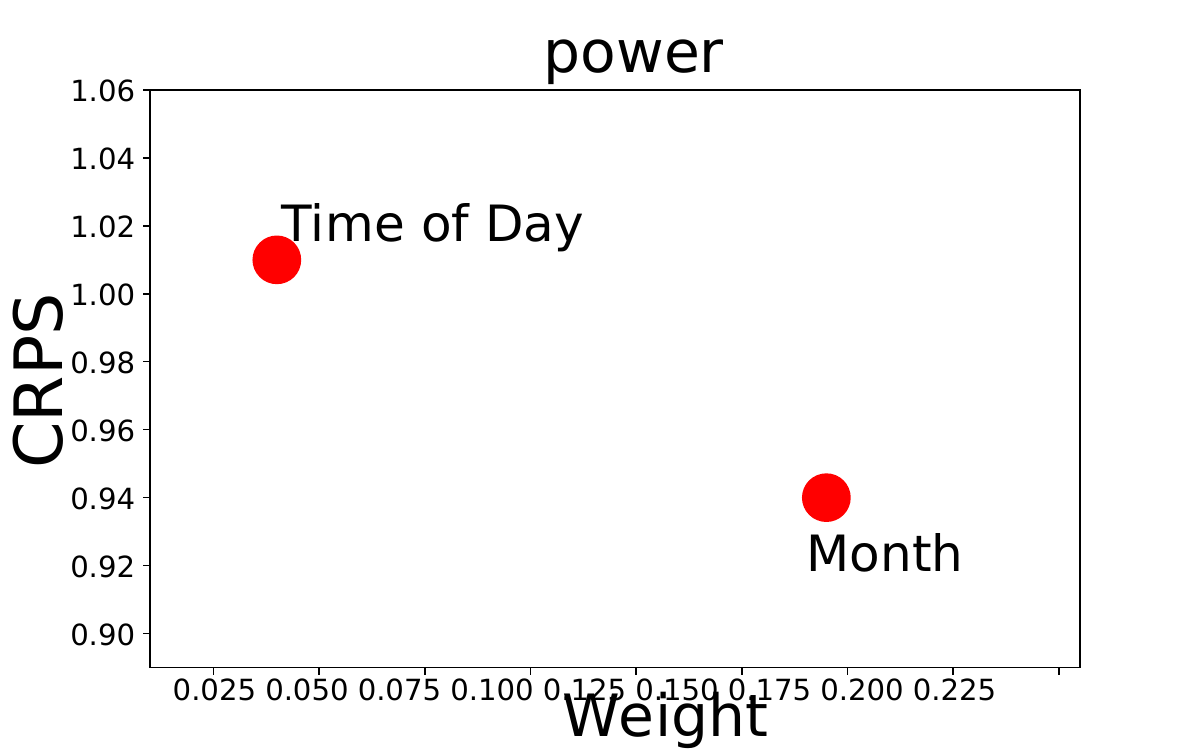}
    \end{subfigure}%
    \begin{subfigure}{.47\linewidth}
    \centering
    \includegraphics[width=.98\linewidth]{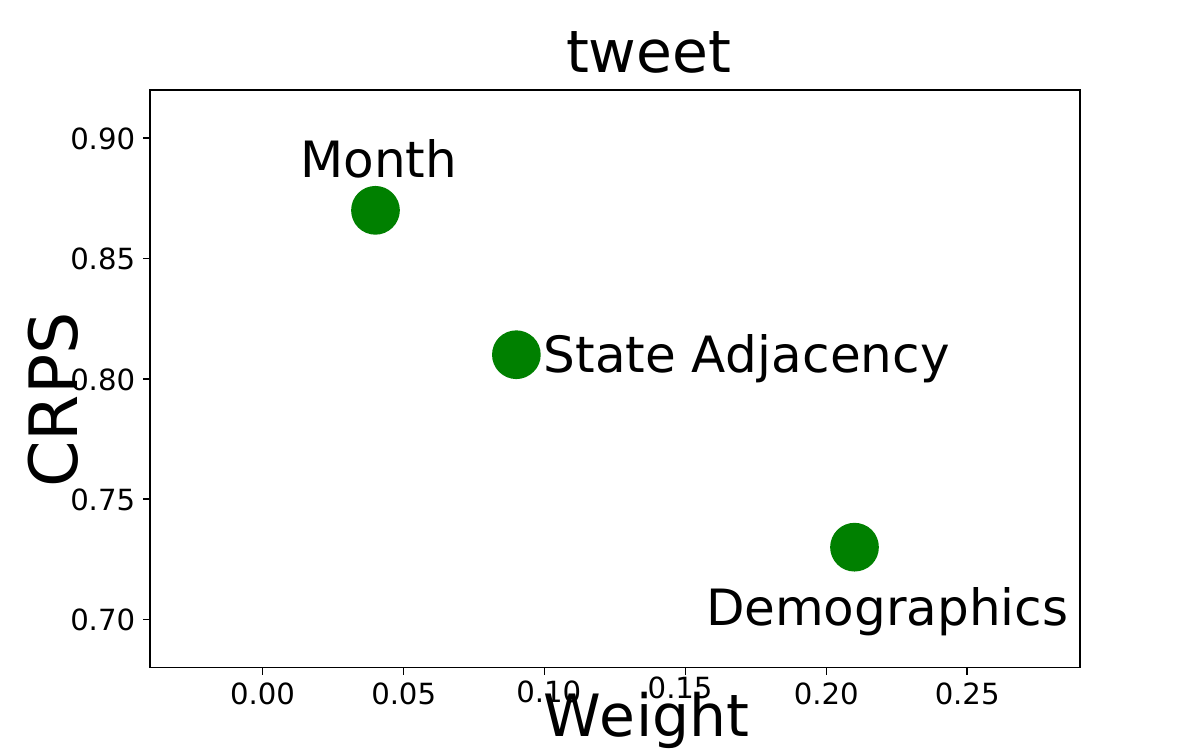}
    \end{subfigure}
    \caption{\textit{CRPS is inversely correlated with attention weights of View selection module.}}
    \label{fig:attencrps}
\end{figure}
\vspace{-.2in}

The attention weights learned by the view selection module in Eqn. \ref{eqn: attenview} informs the learned importance of views that are used for aggregating view-aware latent embeddings. 
We observed that the default view $1$ containing time-series sequences was almost always the most highly weighted views and was necessary for model to perform reasonably. 
Next, we study the correlation between attention weights of other views and the efficacy of the views. We assess the efficacy of a view by training variant of \model with two views: the default view and view we are interested in.
We compare the CRPS scores of these variants to the attention weights (Figure \ref{fig:attencrps}) and see that the CRPS scores are inversely proportional to the view selection module weight.
 This further shows that the view selection module, indeed, selects the most informative views on average to improve the performance of our \model models.

\noindent\textbf{Q5:} \textit{Does the dynamic view-selection module adapt across time to select informative views?}

We provide specific case-studies to show the efficacy of the view selection module in selecting useful views specific to the input sequence by studying the attention weights. 
We describe a case-study related to \texttt{covid19} and one related to \texttt{google-symptoms}.

\noindent\textit{Obs 1: For the \texttt{covid19} task, weights of mobility view are higher than default sequence view during early pandemic (June and July) for many highly populous US states.}

The average attention weight on the default view is over 40\% for all months over all states. However, during the initial 2 months of the dataset (June and July), 12 states observed higher weight for mobility view including populous states such as TX, GA, MA and NY (Supplementary Table \ref{tab:states}).
\model's view selection module adapted by relying more on the mobility view data during the initial months for mortality forecasting.
This concurs with studies that at the initial stages of the pandemic, decrease in mobility was highly correlated with decrease in the disease spread \cite{chang2021mobility}, but later on, this changed~\cite{rodriguez2020deepcovid}.  
In addition, 
line-list data was error-prone during the initial months of the pandemic because the reporting systems were not yet in place \cite{rodriguez_steering_2021}
whereas the mobility features extracted from digital sources were more accurate in real-time \cite{budd2020digital}. 

\noindent\textit{Obs 2:
In relation to \texttt{google-symptoms} task, for the weeks around the peak week, the weights of views other than sequence view decreases by over 30\% on average for all HHS regions.}

The month and HHS adjacency view capture information about seasonal and average region-specific patterns. However, near the peak weeks where wILI values are more volatile, \model relies on the past wILI sequence patterns including values from past weeks for interpolation which is observed by the sudden decrease in attention weights of other views. Therefore, we observed an average decrease of 32.4\% in attention weights of Month and Adjacency view during the 4 weeks around the peak week.
\vspace{-0.15in}
\begin{figure}[h]
    \centering
   \includegraphics[width=.85\linewidth]{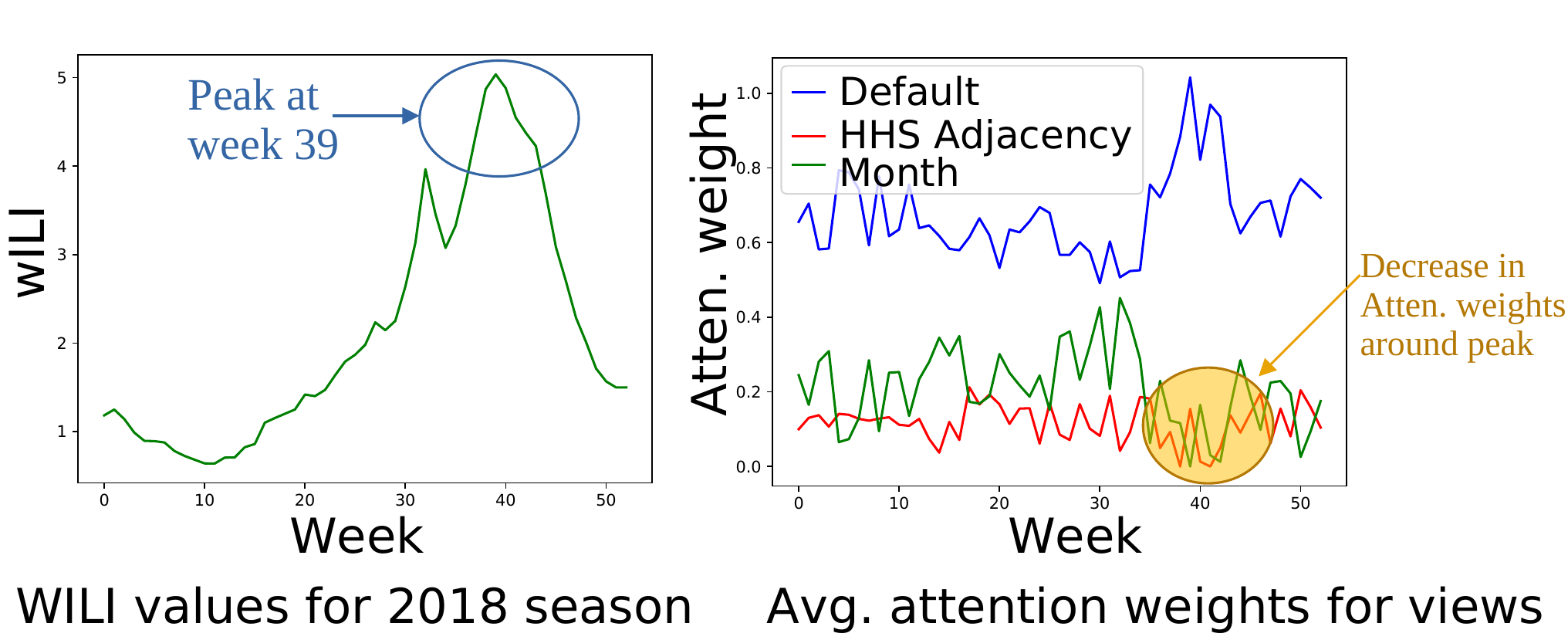}
   \vspace{-0.1in}
    \caption{\textit{Sequence view is more important during peak weeks}}
    \label{fig:2018atten}
\end{figure}
\vspace{-.15in}

An example is shown in Figure \ref{fig:2018atten} for the 2018 flu season. We see that the peak week is at week 40. The weights corresponding the past sequence view sees an increase on weeks 38 - 42 whereas weights of other views decrease.

\vspace{-0.1in}
\section{Conclusion}
We introduced \model, a general non-parametric generative framework for multi-source, multi-modal probabilistic forecasting. 
Our framework successfully tackles the challenge of modeling probabilistic information and uncertainty from diverse data views of multiple modalities across multiple domain benchmarks. 
\model outperformed both state-of-art general probabilistic baselines as well as top domain-specific baselines, resulting in over 25\% improvement in accuracy and calibration metrics. Our case-studies also empirically showed the importance of joint probabilistic modeling and context-sensitive integration from multiple views. 
It automatically adapts to select more useful data views as seen in the \texttt{covid19} and \texttt{google-symptoms} case studies.

\model can easily be applied to any multi-source and multi-modal forecasting task by using appropriate encoders to encode a variety of data where capturing multi-source uncertainty is important.   
Analysis of view selection for specific tasks can also help understand data reliability, the disparity in data quality across sensitive parameters.
As future work, our work can also be extended to tasks such as anomaly detection, change point detection and time-series segmentation where drastic variation in confidence intervals and view selection weights can be a useful predictive indicator of important behaviors.

\par \noindent \textbf{Acknowledgments:} This work was supported in part by the NSF (Expeditions CCF-1918770,
CAREER IIS-2028586, RAPID IIS-2027862, Medium IIS-1955883, Medium IIS-2106961,
CCF-2115126, Small III-2008334), CDC MInD program, ORNL, ONR MURI
N00014-17-1-2656, faculty research awards from Facebook, Google and Amazon and
funds/computing resources from Georgia Tech.

\clearpage
\bibliographystyle{ACM-Reference-Format}
\bibliography{references}
\newpage
\clearpage

\newpage
\clearpage
\appendix
\appendix
\noindent\textbf{\Large Supplementary for the paper "\model: Calibrated and Accurate Multi-view Time-Series Forecasting"}
\vspace{5pt}

We have released the code for \model models and anonymized datasets at: \url{https://github.com/AdityaLab/CAMul}.

\section{Examples of Multi-view Probabilistic Latent Encoders}
We provide examples of Probabilistic Latent Encoders for data views of different modalities that are used in the paper. Due to generality of our framework, variety of neural encoders can be used specific to data view and type of task.

\paragraph{Static features:}
We encode the static features of fixed dimensions of form $R_k^{(j)} \in \mathbf{R}^d$. We employ a feed-forward network to capture the parameters of distribution.

\paragraph{Sequences:}
To capture the latent embedding of sequence $R^{(j)}_k = \{R[0]_k^{(j)}, R[1]_k^{(j)}, \dots, R[T]_k^{(j)}\}$, such as in the default view, we leverage neural sequence encoders  GRU \cite{cho2014learning} as:
\begin{equation}
    \{\tilde{z}[t]_k^{(j)}\}_{t=0}^T =  GRU(\{R[t]_k^{(j)}\}_{t=0}^T)
\end{equation}
where $\{\tilde{z}[t]_k^{(j)}\}_{t=0}^T$ are the intermediate states of GRU.

In order to capture capture long-term relations and prevent over-emphasis on last terms of sequence we employ self-attention layer introduced in \cite{vaswani2017attention}:
\begin{equation}
\begin{split}
    \{\alpha[0]_d, \alpha[1]_d, \dots, \alpha[T]_d\} & = \text{Self-Atten}(\{\tilde{z}[t]_k^{(j)}\}_{t=0}^T)\\
    \mu_k^{(j)}, \sigma_k^{(j)} & = g'_j(\sum_{t=0}^T \alpha[t]_d z[t]_d^{(j)})
\end{split}
\label{eqn:sumall1}
\end{equation}
where $g'_j$ is a single feed-forward layer.

\paragraph{Graph data:}
If the data type of view $j$ contain relations, i.e, the reference sets have an inherent graph structure denoted by $(A^{(j)}, F^{(j)})$, where $A^{(j)} \in \mathbf{R}^{N_j\times N_j}$ is the adjacency matrix (potentially with weights) and $F^{(j)} = \{f_i^{(j)}\}_{i=1}^{N_j}$ are feature vectors of fixed dimensions, then we can use a Graph Neural network architecture \cite{Kipf2017SemiSupervisedCW} to encode the relations in $A^{(j)}$:
\begin{equation}
     \{\tilde{z}_i^{(j)}\}_{i=1}^{N_j} =  GCN(A^{(j)}, \{f_i^{(j)}\}_{i=1}^{N_j})
\end{equation}

Then, we use feed forward layer to derive the distribution parameters:
\begin{equation}
   \mu_i^{(j)}, \sigma_i^{(j)} = g'_j(\tilde{z}_i^{(j)})
\end{equation}

\section{Implementation details}
We used numpy for data processing and PyTorch for model training and inference. We use the properscoring library \cite{properscoring} to implement the CRPS evaluation.

\subsection{Data pre-processing}
\textit{Scaling values:} Since time-series and exogenous features can have wide range of values, we normalize the values of of each features with mean 0 and variance 1. We derive the scaling factors for training dataset and apply the transformation to test set during inference.

\noindent\textit{Chunking time-series for training:} The long training sequences are split using the shingling technique \cite{leskovec2020mining, li2021learning} where we fix a window size $W=10$ and randomly sample chunks of $W$ size from the full sequence over the interval $[1, t-W-1]$ and record the $\tau$ ahead value as ground truth. Note that the reference set of default view $1$ still contains the full length sequence of training set, we only use the split sequence as input for training set.

\subsection{Hyperparameters}
We describe the hyperparameters for \model for all benchmarks. 

\noindent\textit{Multi-view probabilistic encoders}

For the feed-forward networks of static view, we used a 3 layer network with 60 hidden units. We also used 60 hidden units for GRU of sequence views and used a bi-directional version of GRU. For graph views, we used a 2-layer GCN network with 60 hidden units. The final layer outputs 120 dimensional vector which we split into mean and variance. We apply exponentiation on the variance vector to make it positive.
The latent embeddings $z_i^{(j)}$ for all views thus have dimension of 60. 

\noindent\textit{View specific correlation graph}
The networks $l_j^\mu$ and $l_j^\sigma$ is a 2 layer feed forward network with 60 hidden units. Thus, the dimension of view-aware latent embedding is 60.

\noindent\textit{Dynamic view-selection module}
We pass the view-aware latent embeddings and $z_i^{(1)}$ each through a single 60-hidden unit feed forward layer ($h_1$, $h_2$) before computing the cross attention weights. The combined-view embedding $\tilde{u}_i$ is a 60 dimensional vector.

\noindent\textit{Forecast decoder}
We use a 3 layer feed forward layer that inputs the latent embedding of sequence $z_i^{(1)}$ and $\tilde{u}_i$ and has 60 hidden units in each hidden layer with final layer outputting 2 scalars $\mu(y_i), \sigma(y_i)$.
Throughout all layers of \model we apply exponentiation to variance vector/scalar to make it positive. We also use ReLU activation for hidden layers unless specified otherwise. We also use ADAM optimizer \cite{kingma2014adam} for parameter updates.

\paragraph{Note on hyperparameter selection} 
We sampled a validation set containing 10\% of randomly selected chunks of sequences from training set. This was used for model hyperparameter tuning. We describe splitting of dataset into training and test set in Section \ref{sec:bench} below for each benchmark.

We found that the performance was not significantly sensitive to model architecture, batch size or learning rate. We searched over the space of $\{30, 60, 120, 250\}$ for hidden units and mostly optimized for faster convergence. We also searched over $\{10, 20, 50, 80\}$ for batch-size and found $20, 20, 10, 50$ to be most optimal for \texttt{tweet, covid19, google-symptoms, power} respectively. We used $0.005$ as learning rate. We also used early stopping with patience of 150 epochs for faster training.
For \texttt{power} and \texttt{google-symptoms} we observed that \model took less than 700 epochs to converge whereas it took 2000 and 1000 epochs for \texttt{covid19} and \texttt{tweet}. Regarding seed selection, we initialized random seeds to 0 to 19 for numpy and pytorch for the 20 trials done for each benchmarks and didn't observe any significant variation in scores.

\begin{algorithm}[h]
\SetAlgoLined
\SetKwInOut{Input}{Input}
\SetKwInOut{Output}{output}

\Input{Training sequences $\{Y_N\}_{i=1}^N = \{X_i^{(1)}\}_{i=1}^N$, target labels $\{y[t+\tau]_i\}_{i=1}^N$, view data sources $\{X_i^{(j)}\}_{i=1}^N$ for each view $j\in \{2, \dots, K\}$, reference points $\{R_i^{(j)}\}_{i=1}^{N_j}$ for each view $j$}

\For{$i\sim \{1, \dots, N\}$}{

\For{$j \in \{1, \dots, K\}$}{
\tcc{Stochastic embeddings from latent stochastic encoder}
    Sample $z_i^{(j)}$ as Eqn \ref{eqn:latent} using encoder $g_{\theta_j}$\;
    \For{$R_{k}^{(j)} \in \mathcal{R}^{(j)}$}{
        Sample $z_k^{(j)}$ using encoder $g_{\theta_j}$\;
    }
    \tcc{Sample edges for SVCG}
    \For{$R_{k}^{(j)} \in \mathcal{R}^{(j)}$}{
        Add $(k,i)$ to $G^{(j)}$ with probability $k(z_{k}^{(j)}, z_{i}^{(j)})$\;
    }
    Derive $\mu(u_{i}^{(j)}), \sigma(u_{i}^{(j)})$ from sampled edges $\{k: (k,i) \in G^{(j)}\}$ as Eqn \ref{eqn:aggregate}\;
    
    \tcc{Sample from Variational distribution}
    Sample $\hat{u}_{i}^{(j)}$ from variational distribution $q_j$\;
}

\tcc{Aggregate all view-aware embeddings using Dynamic view-selection module}
Compute importance weights $\{\alpha_i^{(j)}\}_{j=1}^K$ using cross-attention as Eqn \ref{eqn: attenview} from $\{\hat{u}_i^{(j)}\}_{j=1}^K$\;
Compute the combined view embedding $\Tilde{u}_{i}^{(j)} = \sum_{j=1}^K \alpha_i^{(j)} \hat{u}_{i}^{(j)}$\;

\tcc{Final output distribution}
Derive $\mu(y_i), \sigma(y_i)$ from $z_{i}^{(1)}$ and $\Tilde{u}_{i}^{(j)}$ via the decoder (Eqn \ref{eqn:decoder})\;

\tcc{Sample ELBO Loss}
Compute $\mathcal{L}_1 = \log P(y[t+\tau]_i | \mu(y_i), \sigma(y_i))$\;
Compute $\mathcal{L}_2 = \sum_{j=1}^K \log P(\hat{u}_i^{(j)} | \mu(u_{i}^{(j)}), \sigma(u_{i}^{(j)})) - \log q_j(\hat{u}_{i}^{(j)} | X_i^{(j)})$\;
Accumulate gradient for loss $\mathcal{L} = -(\mathcal{L}_1 + \mathcal{L}_2)$\;
}
Periodically update the weights of all modules of \model from accumulated gradients using ADAM\;

\caption{Training Algorithm for \model}
\label{alg:training}
\end{algorithm}

\begin{table}[h]
\centering
\scalebox{0.8}{
\begin{tabular}{llrrrrr}
State               & Month & \multicolumn{1}{l}{Seq.} & \multicolumn{1}{l}{Line-list} & \multicolumn{1}{l}{Mobility} & \multicolumn{1}{l}{Adj.} & \multicolumn{1}{l}{Demo.} \\ \hline
\multirow{2}{*}{TX} & June  & 0.32                     & 0.22                          & 0.37                         & 0.03                     & 0.06                      \\
                    & July  & 0.37                     & 0.23                          & 0.31                         & 0.05                     & 0.04                      \\
\multirow{2}{*}{GA} & June  & 0.21                     & 0.34                          & 0.31                         & 0.01                     & 0.13                      \\
                    & July  & 0.24                     & 0.40                          & 0.26                         & 0.02                     & 0.08                      \\
\multirow{2}{*}{MA} & June  & 0.32                     & 0.19                          & 0.36                         & 0.04                     & 0.09                      \\
                    & July  & 0.39                     & 0.29                          & 0.23                         & 0.02                     & 0.07                      \\
\multirow{2}{*}{NY} & June  & 0.25                     & 0.24                          & 0.32                         & 0.04                     & 0.15                      \\
                    & July  & 0.26                     & 0.34                          & 0.27                         & 0.01                     & 0.12                     
\end{tabular}
}
\caption{\textit{Avg. Attention weight of View selection module during June and July for some populous US states.}}
\label{tab:states}
\vspace{-0.1in}
\end{table}

\begin{algorithm}[h]
\SetAlgoLined
\SetKwInOut{Input}{Input}
\SetKwInOut{Output}{output}

\Input{Training sequences $\{Y_N\}_{i=1}^N = \{X_i^{(1)}\}_{i=1}^N$, target labels $\{y[t+\tau]_i\}_{i=1}^N$, view data sources $\{X_i^{(j)}\}_{i=1}^N$ for each view $j\in \{2, \dots, K\}$, reference points $\{R_i^{(j)}\}_{i=1}^{N_j}$ for each view $j$}

\For{$i\sim \{1, \dots, N\}$}{

\For{$j \in \{1, \dots, K\}$}{
\tcc{Stochastic embeddings from latent stochastic encoder}
    Sample $z_i^{(j)}$ as Eqn \ref{eqn:latent} using encoder $g_{\theta_j}$\;
    \For{$R_{k}^{(j)} \in \mathcal{R}^{(j)}$}{
        Sample $z_k^{(j)}$ using encoder $g_{\theta_j}$\;
    }
    \tcc{Sample edges for SVCG}
    \For{$R_{k}^{(j)} \in \mathcal{R}^{(j)}$}{
        Add $(k,i)$ to $G^{(j)}$ with probability $k(z_{k}^{(j)}, z_{i}^{(j)})$\;
    }
    Derive $\mu(u_{i}^{(j)}), \sigma(u_{i}^{(j)})$ from sampled edges $\{k: (k,i) \in G^{(j)}\}$ as Eqn \ref{eqn:aggregate}\;
    
    \tcc{Sample from Variational distribution}
    Sample $\hat{u}_{i}^{(j)}$ from variational distribution $q_j$\;
}

\tcc{Aggregate all view-aware embeddings using Dynamic view-selection module}
Compute importance weights $\{\alpha_i^{(j)}\}_{j=1}^K$ using cross-attention as Eqn \ref{eqn: attenview} from $\{\hat{u}_i^{(j)}\}_{j=1}^K$\;
Compute the combined view embedding $\Tilde{u}_{i}^{(j)} = \sum_{j=1}^K \alpha_i^{(j)} \hat{u}_{i}^{(j)}$\;

\tcc{Final output distribution}
Derive $\mu(y_i), \sigma(y_i)$ from $z_{i}^{(1)}$ and $\Tilde{u}_{i}^{(j)}$ via the decoder (Eqn \ref{eqn:decoder})\;

\tcc{Sample ELBO Loss}
Compute $\mathcal{L}_1 = \log P(y[t+\tau]_i | \mu(y_i), \sigma(y_i))$\;
Compute $\mathcal{L}_2 = \sum_{j=1}^K \log P(\hat{u}_i^{(j)} | \mu(u_{i}^{(j)}), \sigma(u_{i}^{(j)})) - \log q_j(\hat{u}_{i}^{(j)} | X_i^{(j)})$\;
Accumulate gradient for loss $\mathcal{L} = -(\mathcal{L}_1 + \mathcal{L}_2)$\;
}
Periodically update the weights of all modules of \model from accumulated gradients using ADAM\;

\caption{Training Algorithm for \model}
\label{alg:training}
\end{algorithm}

\begin{table}[h]
\centering
\scalebox{0.9}{
\begin{subtable}{.45\textwidth}
\caption{\texttt{power}}
\centering
\begin{tabular}{lccc|ccc}
\hline
            & \multicolumn{3}{c|}{Isotonic}                 & \multicolumn{3}{c}{DC}                        \\
Model       & CRPS          & CS            & IS            & CRPS          & CS            & IS            \\ \hline
SARIMA      & 1.32          & 0.14          & 3.28          & 1.7           & 0.23          & 3.95          \\
DeepAR      & 0.67          & 0.04          & 1.55          & 0.71          & 0.08          & 1.64          \\
DSSM        & 0.62          & 0.04          & 1.62          & 0.81          & 0.09          & 1.84          \\
RNP         & 0.87          & 0.17          & 2.55          & 0.93          & 0.16          & 2.58          \\
GP          & 1.03          & 0.12          & 2.78          & 1.03          & 0.12          & 2.78          \\
GraphDF-RBF & 0.92          & 0.13          & 1.45          & 1.04          & 0.18          & 1.31          \\ \hline
\model       & \textbf{0.43} & \textbf{0.02} & \textbf{0.91} & \textbf{0.47} & \textbf{0.04} & \textbf{1.02}
\end{tabular}
\end{subtable}
}

\scalebox{0.9}{
\begin{subtable}{.45\textwidth}
\caption{\texttt{tweet}}
\begin{tabular}{lccc|ccc}
\hline
            & \multicolumn{3}{c|}{Isotonic}                 & \multicolumn{3}{c}{DC}                        \\
Model       & CRPS          & CS            & IS            & CRPS          & CS            & IS            \\ \hline
SARIMA      & 1.18          & 0.22          & 3.39          & 1.27          & 0.33          & 3.25          \\
DeepAR      & 1.02          & 0.18          & 1.24          & 1.17          & 1.15          & 1.36          \\
DSSM        & 1.19          & 0.08          & 1.15          & 1.21          & 0.17          & 1.83          \\
RNP         & 1.09          & 0.15          & 1.89          & 1.17          & 0.13          & 2.05          \\
GP          & 1.25          & 0.15          & 1.86          & 1.27          & 0.15          & 2.38          \\
GraphDF-RBF & 0.7           & 0.1           & 1.61          & 0.75          & 0.08          & 1.92          \\
GraphDF-Adj & 1.15          & 0.18          & 2.63          & 1.17          & 0.18          & 2.39          \\ \hline
\model       & \textbf{0.55} & \textbf{0.06} & \textbf{0.68} & \textbf{0.67} & \textbf{0.07} & \textbf{0.86}
\end{tabular}
\end{subtable}
}

\scalebox{0.9}{
\begin{subtable}{.45\textwidth}
\caption{\texttt{covid19}}
\begin{tabular}{lccc|ccc}
\hline
            & \multicolumn{3}{c|}{Isotonic}                 & \multicolumn{3}{c}{DC}                        \\
Model       & CRPS          & CS            & IS            & CRPS          & CS            & IS            \\ \hline
SARIMA      & 110.4         & 0.34          & 8.66          & 106.0         & 0.39          & 8.93          \\
DeepAR      & 57.9          & 0.16          & 3.72          & 58.2          & 0.21          & 3.80          \\
DSSM        & 84.1          & 0.2           & 3.08          & 91.9          & 0.28          & 4.41          \\
RNP         & 74.7          & 0.31          & 5.52          & 67.1          & 0.34          & 7.30          \\
GP          & 44.0          & 0.26          & 5.45          & 40.4          & 0.29          & 5.63          \\
GraphDF-RBF & 68.7          & 0.23          & 5.25          & 62.8          & 0.27          & 5.54          \\
GraphDF-Adj & 73.8          & 0.29          & 4.10          & 65.8          & 0.29          & 4.12          \\ \hline
DeepCovid   & 55.9          & 0.16          & 4.40          & 49.7          & 0.16          & 4.34          \\
CMU-TS      & 40.9          & 0.13          & 4.81          & 36.0          & 0.13          & 5.23          \\ \hline
\model       & \textbf{27.6} & \textbf{0.12} & \textbf{2.28} & \textbf{23.7} & \textbf{0.12} & \textbf{2.20}
\end{tabular}
\end{subtable}
}

\scalebox{0.9}{
\begin{subtable}{.45\textwidth}
\caption{\texttt{google-symptoms}}
\begin{tabular}{lccc|ccc}
\hline
            & \multicolumn{3}{c|}{Isotonic}                 & \multicolumn{3}{c}{DC}                        \\
Model       & CRPS          & CS            & IS            & CRPS          & CS            & IS            \\ \hline
SARIMA      & 1.12          & 0.42          & 2.63          & 1.28          & 0.35          & 2.71          \\
DeepAR      & 0.81          & 0.15          & 1.53          & 0.89          & 0.14          & 1.63          \\
DSSM        & 0.72          & 0.15          & 2.30          & 0.85          & 0.14          & 2.44          \\
RNP         & 0.89          & 0.23          & 2.42          & 0.93          & 0.27          & 1.73          \\
GP          & 0.86          & 0.10          & 1.90          & 0.99          & 0.1           & 1.92          \\
GraphDF-RBF & 0.78          & 0.11          & 1.21          & 0.92          & 0.12          & 1.32          \\
GraphDF-Adj & 0.81          & 0.09          & 2.98          & 0.96          & 0.1           & 2.81          \\ \hline
EpiDeep     & 1.53          & 0.12          & 2.62          & 1.15          & 0.14          & 2.19          \\
EpiFNP      & 0.57          & 0.11          & 0.59          & 0.65          & 0.07          & 0.64          \\ \hline
\model       & \textbf{0.45} & \textbf{0.05} & \textbf{0.52} & \textbf{0.49} & \textbf{0.06} & \textbf{0.56}
\end{tabular}
\end{subtable}
}

\caption{\textit{Evaluation scores of \model and baselines after applying post-hoc calibration. We use Isotonic regression \cite{kuleshov2018accurate} and Distribution calibration \cite{song2019distribution} methods. The evaluation scores of \model is  statistically significantly better and does not change much due to post-hoc methods, implying that our approach produces well-calibrated forecasts without need for such post-hoc correction methods.}}

\label{tab:posthoc}
\end{table}

\end{document}